\documentclass{article}

\usepackage{arxiv}

\usepackage[utf8]{inputenc} 
\usepackage[T1]{fontenc}    
\usepackage{hyperref}       
\usepackage{url}            
\usepackage{booktabs}       
\usepackage{amsfonts}       
\usepackage{nicefrac}       
\usepackage{microtype}      
\usepackage{lipsum,bm}
\usepackage{graphicx}
\graphicspath{ {./images/} }

\newcommand{\mb}{\mathbb} 
\newcommand{\vc}{\mathbf} 
\newcommand{\gvc}{\bm}

\usepackage{amsmath}
\newcommand{\mc}{\mathcal}

\title{A generalized kernel machine approach to  identify  higher-order composite effects  in multi-view datasets}
\author{
 Md Ashad Alam\\
  Tulane Center of Bioinformatics and Genomics\\
  Department of  Biostatistics and Data Science\\
  Tulane University\\
  New Orleans, LA 70112, USA\\
   \And
Chuan Qiu \\
  Tulane Center of Bioinformatics and Genomics\\
  Department of  Biostatistics and Data Science\\
  Tulane University\\
  New Orleans, LA 70112, USA\\
  \And
 Hui Shen \\
   Tulane Center of Bioinformatics and Genomics\\
  Department of  Biostatistics and Data Science\\
  Tulane University\\
  New Orleans, LA 70112, USA\\
     \And
 Yu-Ping Wang\\
   Tulane Center of Bioinformatics and Genomics\\
  Department of Biomedical Engineering\\
  Tulane University\\
  New Orleans, LA 70118, USA
    \And
Hong-Wen Deng\\
   Tulane Center of Bioinformatics and Genomics\\
  Department of  Biostatistics and Data Science\\
  Tulane University\\
  New Orleans, LA 70112, USA\\
}

\begin{document}
\maketitle
\begin{abstract}
In recent years, comprehensive study of multi-view datasets (e.g., multi-omics and imaging scans) has been a focus and forefront in biomedical research. State-of-the-art biomedical technologies are enabling us to collect multi-view biomedical datasets for the study of complex diseases. While all the views of data tend to explore complementary information of a disease, multi-view data analysis with complex interactions is challenging for a deeper and holistic understanding of biological systems. In this paper, we propose a novel generalized kernel machine approach to identify higher-order composite effects in multi-view biomedical datasets. This generalized semi-parametric (a mixed-effect linear model) approach includes the  marginal and joint Hadamard product of features from different views of data. The proposed kernel machine approach considers multi-view data as predictor variables to allow more thorough and comprehensive modeling of a complex trait. The proposed method can be applied to the study of any disease model, where multi-view datasets are available. We  applied our approach to both synthesized datasets and real multi-view datasets from  adolescence brain development and osteoporosis study, including an imaging scan dataset and  five omics  datasets. Our experiments demonstrate that the proposed method can effectively identify higher order composite effects and suggest that corresponding features (genes, region of interests, and chemical taxonomies) function in a concerted effort. We show that the proposed method is more generalizable than existing ones.
\end{abstract}


\section{Introduction}
\label{Sec:int}
\begin{figure}
\begin{center}
\includegraphics[width=8cm, height=5cm]{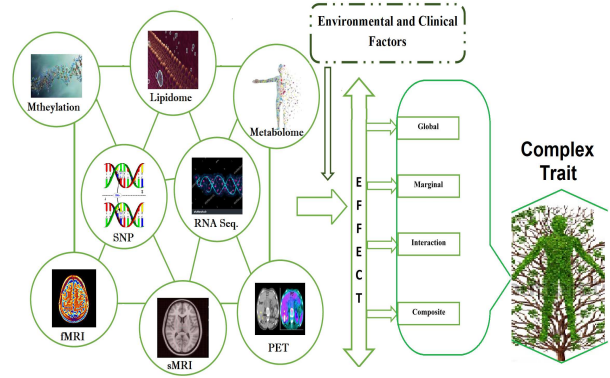}
\caption{An illustration of  different omics data, imaging scans,  and environment factors along  with their  global, marginal, interaction and composite on a complex trait.}
\label{fig1:mvd}
\end{center}
\end{figure}

Modern technology allows us to overcome the challenges of assembling a multi-view dataset with a powerful combination of flexibility and low cost. Human complex diseases are typically expressed by the aberrant interplay of multiple biological sources and environmental factors. While single biomedical data view such as genome (whole genome sequencing), transcriptome (RNA sequencing), epigenome (e.g., DNA methylation), proteome, metabolome, lipidome, and medical images offers an essential framework for detecting biological variants contributing to complex diseases, an in-depth understanding of biological mechanisms by examining only single data may not suffice. Most biomedical data analysis methods are applicable to single or dual-views of datasets, without making full use of multi-view datasets. In addition, multi-view biomedical datasets are often complementary to each other and correspond to different feature spaces \cite{Canzler-20, Ma-19, Kim-18, Rappoport-18, Hasin-17}. Thus, their comprehensive and systematic analysis has the potential to uncover unknown biological mechanisms of complex  diseases (Figure \ref{fig1:mvd}).  With the advent of recent technologies that make assembling of multi-view biomedical datasets (e.g., multi-omics and imaging scans) of a complex diseases possible, development of efficient analytic frameworks for such datasets is becoming an active area of scientific inquiry \cite{Ma-19b, Rattray-18, Hung-17, Yang-16, Schmidt-14}. For an effective diagnosis of a  complex disease,  different types of medical imaging techniques,  including computer tomography (CT), ultrasound, functional magnetic resonance imaging (fMRI), structural MRI (sMRI),  positron emission tomography (PET) scans, diffusion tensor imaging (DTI), and X-Ray are used  \cite{ Sui-20, Alam-18c, Cabral-17, Hansen-15, Deco-15}.

The high demand for characterizing complex diseases in biomedical science and the available technologies justify the need for more effective biomedical data integration approaches.  Linear data integration approaches including supervised,  unsupervised, semi-supervised learning, multi-omics factor analysis, and highlight matrix factorization methods are used  to gain a  deeper,  more holistic understanding of the biological mechanism for complex diseases \cite{Sathyanarayanan-19, Ma-19, Argelaguet-18, Ritchie-15,  Hastei-book}. These are widely used and  validated approaches  but they perform poorly when data structure is non-linear and data comes from a multi-modal distribution \cite{Hofmann-08}. As a consequence, non-linear  integrated approaches (e.g., kernel based machine)  are an important feature   in comprehensive analysis of  multi-view biomedical datasets \cite{Nascimento-16, Borgwardt-05, Lanckriet-04}. The positive definite kernel based machine approach  can overcome the non-linearity problem  as well as  that of  dimensionality of multi-view biomedical datasets  \cite{Yan-17, Zheng-16, Ashad-14T, Alam-16a, Schlkof-book}.  This approach can be useful in effective integration of multi-view biomedical data.

In this study,  we  introduce  a generalized kernel machine approach  to identify higher-order composite effects (GKMAHCE) in multi-view datasets. While several kernel machine  approaches have been employed to identify marginal, interaction effects of datasets  with  continuous  data,  the underlying challenge remains of how to make a generalized kernel machine based model to estimate marginal, interaction, and composite effects of multi-view  dataset with categorical  data, which is often the case in  complex  diseases \cite{ Sui-18, Bersanelli-16,  Zhao-15, Ionita-13, Camps-07}. This proposed approach considers multi-view data as predictor variables to allow more thorough and comprehensive modelling of complex traits.

Here, we compared the performance of our proposed method with that of existing methods using  synthesized datasets and real multi-view adolescence brain development and  osteoporosis  datasets, with brain imaging and five- omics data. To realize adolescence brain development, sex differences in human brain are essential to understand their anatomical foundations in the brain with three  view datasets:  genomic,  non-genomic  and  fMRI  datasets \cite{Satterthwaite-46}.   Osteoporosis is a bone disorder,  which results from a loss of bone mineral density (BMD) \cite{Sozen-17}. For more information of these data, we refer the reader to  Section $4.2$. Within each dataset, this study explores novel genome, epigenome, transcriptome, and chemical mechanisms, as well as robustly and efficiently identifies  corresponding  factors.  To  validate  the  results,  we  performed
Kyoto Encyclopedia of Genes and Genomes (KEGG) pathway analysis, the gene ontology analysis of biological process category  and  gene-gene interaction  networks  analysis.  We  also  draw  causal  connections,  gene- gene associations, and gene-chemical associations for both of the datasets. This paper is highly novel in  the following aspects:

\begin{itemize}
    \item  To the best of our knowledge, we address  a  generalized kernel machine approach to  identify  higher-order composite effects.
    \item   For statistical testing of the marginal, the interaction, and the composite effects,  we derive the test  statistics.
    \item To show that the proposed approach is highly efficient, we experiment  with  synthesized datasets and three-view  of  brain development in adolescence and  five-view of  osteoporosis datasets. As far as we know,  this is the first five omics integration study.
    \item  Finally, we perform  a comprehensive  pathway and network-based analysis for the biological validation of  the selected genes.
    \end{itemize}

The  rest of this paper is organized as  follows. In Section 2, we  discuss some contemporary  related work. In Section $3$, we explain  the proposed approach,  generalized kernel machine approach to  identify  higher-order composite effects  in multi-view datasets and  test statistics for  statistical testing for higher order  composite  effects..  In Section $4$, we describe the experiments conducted on both synthesized datasets and two  real datasets:  imaging  and genetics for adolescence brain development and  multi-omics datasets of  osteoporosis studies. We conclude the paper with a discussion on major findings and future research in Section $5$. Detailed derivation  of the proposed method along with  Satterthwaite approximation to the score test and the applications to  real  dataset can be found in the supplementary materials.

\section{Related work}
Kernel based multi-view data integration  approaches allow the joint analysis of multiple  data types to provide a global view of the biological functions and offer insights into the nature of the interactions between the different datasets.  These methods offer useful ways to learn how an extensive collection of genetic variants are associated with complex phenotypes and can be used to explore the relationship between genetic markers and a disease state  \cite{Camps-07, Yu-11, Kung-14}.

The  marginal, interaction, and composite effect identifications  are becoming a common challenge to multidimensional imaging and multi-omics data analysis.  Preliminary  works in kernel machine methods have been boldly pursued by \cite{Liu-07}, where a single modal dataset was used to test for a genetic pathway effect.  \cite{Li-12} also proposed a kernel machine-based method for gene-gene interactions. They treated each gene as a testing unit for gene-gene interactions. A kernel machine method was then proposed for detecting multiple factor interactions where a smoothing spline-ANOVA decomposition method was adopted \cite{Li-12}. However, these approaches only use single or pairwise datasets. A sequence kernel association tests (SKAT) for genome wide association studies is introduced in \cite{Ionita-13}. While SKAAT is widely used in genomics studies, it is limited to only  a single omics. In addition,  \cite{Ge-15} have  proposed a kernel machine method for detecting the effects of interactions between multidimensional variable sets. This is an extended model of  \cite{Li-12} to jointly model genetic and non-genetic features and their interactions. A microbiome regression based kernel association test (MiRKAT) has been proposed by \cite{Zhao-15}, which directly regresses the outcome on a single omics dataset, microbiome profiles, through the semi-parametric kernel machine regression framework.

Recently,  \cite{Alam-18c} has proposed a kernel machine method for detecting higher-order interactions in three views datasets which was applied to schizophrenia  with a continuous trait . Although a generalized  version of such methods for more than three datasets as well as both categorical and continuous traits remains elusive, such a method for multi-view biomedical data analysis is necessary to discover new information about biological systems and complex diseases.

\section{Approach}
\label{Sec:med}
 Suppose we have $n$ independent identical distributed (IID) subjects $y_i$ $(i = 1, 2, \cdots, n)$ with $(q-1)$ covariates $X_i= [X_{i1}, X_{i2}\cdots X_{iq}]^T$ and m-view datasets, $\vc{M}_{i}^{(1)}, \cdots, \vc{M}_{i}^{(m)}$. Assuming that  $y_i$ follows a distribution in the exponential family with density
\begin{eqnarray}
\label{efe1}
f(y_i, \theta_i, \gamma)= exp\left\{ \frac{y_i\theta_i- c_1(\theta_i)}{\gamma/w_i} + c_2(y_i, \gamma)\right\},
\end{eqnarray}
where $\theta_i$ and  $\gamma$ are the location and scale parameters,  $c_1 (\cdot)$ and $c_2 (\cdot)$ are known functions, and $w_i$ is know weight, respectively. The mean and variance of $y_i$ satisfy  $ \rm{E}(y_i)= \frac{\partial c_1(\theta_i)}{\partial\theta_i}= c_1^\prime(\theta_i)$ and $\rm{Var}(y_i)= \gamma w_i = c_1^{\prime\prime}(\theta_i)$.  In the following generalized semiparametric model,  we associate the output $y_i$ with covariates including intercept and $m$-view datasets:
\begin{eqnarray}
\label{me1}
g(y_i)= X_i^T\vc{\gvc{\beta}} + f(\vc{M}_{i}^{(1)}, \cdots, \vc{M}_{i}^{(m)}),
\end{eqnarray}
 where  $g(\cdot)$ is a known monotone link function,  $X_i$ is a $q\times 1$ vector of covariates including the intercept for the $i-$th subject,  $\gvc{\beta}$ is  a $q\times 1$ vector of fixed effects,  $f$ is an unknown function on the product domain, $\mc{M}=\mc{M}^{(1)} \otimes \mc{M}^{(2)}\otimes, \cdots, \otimes \mc{M}^{(m)}$ with  $\vc{M}_{i}^{(\ell)} \in \mc{M}_\ell, \ell= 1, 2, \cdots m$.  
 
  \begin{table*}
 \begin{center}
\caption {Family and link functions of generalized liner model}
\label{tb:pathwayall}
 \begin{tabular}{ccccccccccc} \hline
\rm{Error family} &\rm{Link}&\rm{Inverse of link}&\rm{Use}\\\hline
\rm{Gaussian} &\rm{Identity,} g(y)=y& $y=g^{-1}(y)$ & \rm{Normally distributed  set of data},$[-\infty, \infty)$\\\hline
\rm{Poisson} &\rm{Log,} g(y)=log(y)& \rm{Exponential, exp(y)}& \rm{equidispersed count set of data}\\\hline
\rm{Binomial} &\rm{Logit,} $g(y)=log(\frac{y}{1-y})$& $\frac{exp(y)}{1+exp(y)}$& \rm{ Binary  set of data}, 0/1\\\hline
\rm{Gamma} &\rm{Inverse gamma}& $\frac{1}{y}$& \rm{ Positive continuous   set of data}, $(0, \infty)$\\\hline
\end{tabular}	
\end{center}
\end{table*} 	
According to the ANOVA decomposition, the function, $f$ can be extended  as:
\begin{multline}
\label{me2}
f(\vc{M}^{(1)}_{i}, \cdots, \vc{M}^{(m)}_{i}) =\\ \sum_{\ell=1}^m h_{\vc{M}^{(\ell)}}(\vc{M}_{i}^{(\ell)}) +  \sum_{\ell < \xi}h_{\vc{M}^{(\ell)}, \vc{M}^{(\xi)}} (\vc{M}^{(\ell)}_{i}, \vc{M}^{(\xi)}_{i}) +\\ \sum_{\ell < \xi< \zeta}h_{\vc{M}^{(\ell)}\times \vc{M}^{(\xi)}\times \vc{M}^{(\zeta)}} (\vc{M}^{(\ell)}_{i}, \vc{M}^{(\xi)}_{i},  \vc{M}^{(\zeta)}_{i}) + \cdots \\+ h_{\vc{M}^{(1)}\times \vc{M}^{(2)}\times \vc{M}^{(3)} \times \cdots, \times \vc{M}^{(m)}} (\vc{M}^{(1)}_{i}, \vc{M}^{(2)}_{i}, \cdots, \vc{M}^{(m)}_{i}), 
\end{multline}
where,  $h_{\vc{M}^{(\ell)}} (\vc{M_i}^{(\ell)})$'s are the  main effects of the respective dataset ($\ell: 1, 2, \cdots m$),      $h_{{\vc{M}^{(\ell)}}, {\vc{M}^{(\xi)}}}(\vc{M}_i^{(\ell)},\vc{M}_i^{(\xi)})$'s  are pairwise interactions effects, $h_{{\vc{M}^{(\ell)}}\times {\vc{M}^{(\xi)}}, {\vc{M}^{(\zeta)}}}(\vc{M}_i^{(\ell_1)},\vc{M}_i^{(\ell)},\vc{M}_i^{(\zeta)})$'s  are the interactions effects of three datasets and so on.  The functional space, RKHS, decomposes as: 
\begin{multline}
\label{me3}
 \mc{H}= \mc{H}_{\vc{M}^{(1)}}\oplus\mc{H}_{\vc{M}^{(2)}} \oplus\cdots\oplus\mc{H}_{\vc{M}^{(m)}}\oplus\mc{H}_{\vc{M}^{(1)}\times \vc{M}^{(2)}}\oplus\mc{H}_{\vc{M}^{(1)}\times \vc{M}^{(3)}} \\\oplus\cdots\oplus\mc{H}_{\vc{M}^{(1)}\times \vc{M}^{(m)}}\oplus\mc{H}_{\vc{M}^{(2)}\times \vc{M}^{(3)}} \\ \oplus\cdots\oplus \mc{H}_{\vc{M}^{(1)}\times \vc{M}^{(2)}\times \vc{M}^{(3)}}  \oplus\cdots\oplus \mc{H}_{\vc{M}^{(1)}\times \vc{M}^{(2)}\times\cdots \times \vc{M}^{(m)}},
\end{multline}
 equipped with an inner product, $\langle \cdot, \cdot\rangle$ and a norm $\|\cdot\|_\mc{H}.$ 
 This  generalizes kernel regression by allowing the model to be related to the response variable via a link function including kernel regression, logistics kernel  regression, and Poisson kernel  regression. For the binary data,  the link function  $g(p)= \rm{logit} (p)=log\frac{p}{1-p}$ provides the logistic kernel regression;  for the count data,    $g(p)= log(p)$ provides the Poisson kernel regression; for  Gaussian  data,  $g(p)= p$ gives  classical kernel regression.
The main  aspect  of this paper is applying this to   five views: genome, epigenome, transcriptome,  metabolome, and lipidome along with Low BMD and High BMD  information of the subject. To do this,  assume that  we have   $n$ IID subjects under investigation; $y_i\, (i =1, 2, \cdots n)$ is a  binary  (LBMD or HBMD) phenotype for the $i$-th subject.  We  associate the clinical covariates (e.g., age,  weight, height) with five views including genome, epigenome, transcriptome,  metabolome, and lipidome.

\par Let  $\vc{X}_i$ denoted the  $(q-1)$ covariates, where $X_{ij}, j=1, 2, \cdots (q-1)$ is a measure of  the $i$-th subject.
  
  Let   $M_i^{(1)}=[M_{i1}^{(1)},  \cdots,  M_{is}^{(1)}]$,  $M_i^{(2)}=[M_{i1}^{(2)},  \cdots,  M_{id}^{(2)}]$, $M_i^{(3)}=[M_{i1}^{(3)}] \cdots,  M_{id}^{(3)}] $,   $M_i^{(4)}=[M_{i1}^{(4)},  \cdots,  M_{ic}^{(4)}]$, and $M_i^{(5)}=[M_{i1}^{(5)}, \cdots,  M_{il}^{(5)}]$,  
  are a  gene with $s$ SNP markers, a gene  with  $d$ methylation profiles,  a gene  with  RNA-Seq,  a chemical taxonomy class with $c$ metabolite  profiles, and  a chemical class with $l$ lipid  profiles   of the $i$-th subject, respectively. Under this setting,  Eq. (\ref{me1}), Eq. (\ref{me2}) and Eq. (\ref{me3}) becomes:
\begin{eqnarray}
\label{me4}
\rm{logit}(p_i)= X_i^T\vc{\gvc{\beta}} + f(\vc{M}_{i}^{(1)},\vc{M}_{i}^{(2)}, \vc{M}_{i}^{(3)},\vc{M}_{i}^{(4)},\vc{M}_{i}^{(5)}),
\end{eqnarray}
where  $\rm{logit}(p_i)= \rm{Pr} \left( y_i=1 |  X_i, \vc{M}_{i}^{(1)},\vc{M}_{i}^{(2)}, \vc{M}_{i}^{(3)},\vc{M}_{i}^{(4)},\vc{M}_{i}^{(5)}\right)$.\vspace{2mm}

\subsection{Model estimation}
We can estimate the function $f\in \mc{H}$ by minimizing the  penalized  loss function of Eq. (\ref{me4}) as:
\begin{eqnarray}
\label{mee1}
\mc{L}(\vc{y},\gvc{\beta}, f)= \frac{1}{n}\sum_{i=1}^n L\left (y_i,X_i^T\gvc{\beta}+F)\right)+\frac{\lambda}{2}\mc{J}(f)
\end{eqnarray}
where L is a loss function,  $F=f(\vc{M}_{i}^{(1)},\vc{M}_{i}^{(2)}, \vc{M}_{i}^{(3)}, \vc{M}_{i}^{(4)}, \vc{M}_{i}^{(5)})$  $\mc{J}(\cdot)= \|\cdot\|_\mc{H}^2$ is a roughness penalty  with tuning parameter $\lambda$. By considering  the negative log-likelihood of the binomial distribution,
\[ L= In ( 1+ e^{-y_i (X_i^T\gvc{\beta}+F)}),\]
the problem becomes as a kernel logistic regression problem:
\begin{eqnarray}
\label{mee1a}
\mc{L}(\vc{y},\gvc{\beta}, f)= \frac{1}{n}\sum_{i=1}^n In\left (1+e^{-y_i (X_i^T\gvc{\beta}+F)}\right)+\frac{\lambda}{2}\mc{J}(f)
\end{eqnarray}
\vspace{2mm}
 It is known that the complete function space of  Eq. (\ref{me2}), $\mc{H}$, has the orthogonal decomposition. Hence the function  $\mc{J}(\cdot)$ can be decomposed accordingly (see the supplementary material for details).
\par It is  convenient and mathematically friendly to  minimize  numerically.  Let $p_i$  be the probability of the $i^{th}$  observation and $\vc{p}= [p_1, p_2, p_3,\cdots, p_n]^T$

 Following many in the literature,  we have a liner mixed effects model  for  $\vc{p}$  such that
\begin{eqnarray}
\label{mee8}
logit (\vc{p})&=&\vc{X}\gvc{\beta}+\vc{h}_{\vc{M}^{(1)}}+\vc{h}_{\vc{M}^{(2)}}+\vc{h}_{\vc{M}^{(3)}} \nonumber\\&+& \vc{h}_{\vc{M}^{(1)}\times \vc{M}^{(2)}} + \vc{h}_{\vc{M}^{(1)}\times \vc{M}^{(3)}}+ \vc{h}_{\vc{M}^{(1)}\times \vc{M}^{(4)}}\nonumber\\&+&\vc{h}_{\vc{M}^{(1)}\times \vc{M}^{(5)}}+\vc{h}_{\vc{M}^{(2)}\times \vc{M}^{(3)}}+ \vc{h}_{\vc{M}^{(2)}\times \vc{M}^{(4)}} \nonumber\\&+& \vc{h}_{\vc{M}^{(2)}\times \vc{M}^{(5)}}+ \vc{h}_{\vc{M}^{(1)}\times \vc{M}^{(2)}\times \vc{M}^{(3)}}+ \cdots\nonumber\\&\vdots&
 \nonumber\\&+&
\vc{h}_{\vc{M}^{(1)}\times \vc{M}^{(2)}\times \vc{M}^{(3)} \times \vc{M}^{(4)} \times \vc{M}^{(5)}},
\end{eqnarray}
where $\gvc{\beta}$ is a coefficient vector of fixed effects, 
$\vc{h}_{\vc{M}^{(1)}}$, $\vc{h}_{\vc{M}^{(2)}}$, $\vc{h}_{\vc{M}^{(3)}}$,  $\vc{h}_{\vc{M}^{(1)}\times \vc{M}^{(2)}}$,  $\vc{h}_{\vc{M}^{(1)}\times \vc{M}^{(3)}}$,  $\vc{h}_{\vc{M}^{(2)}\times \vc{M}^{(3)}}$ and $\vc{h}_{\vc{M}^{(1)}\times \vc{M}^{(2)}\times \vc{M}^{(3)}}$ are independent random effects with normal distributions. The relationship of Eq.  (\ref{mee1})  and Eq. (\ref{mee8})  confirm that all of the  effects obtained by optimizing the loss function in Eq. (\ref{mee1}), are  the equal  to  the best linear unbiased predictors (BLUPs) of the linear mixed effects model in Eq. (\ref{mee8}). This relationship makes   a accessibility for  testing the  variance component instead of testing  the nonparametric function in hull hypothesis, which we can estimate  using the restricted maximum likelihood (ReML) approach (see the supplementary material for details).

\subsection{Statistical testing}
\label{Sec:test}
In this section, we address the test statistic of the overall effect, marginal effects,   different interaction effects, and  different composite effects.
\subsubsection{{\bf Overall hypothesis testing}}
According to our model, the testing overall effect is
 \[H_0:  h_{\vc{M}^{(1)}}= \cdots  =  h_{\vc{M}^{(1)}\times \vc{M}^{(2)}}= \cdots =   \cdots \]
 \[= h_{\vc{M}^{(1)}\times \vc{M}^{(2)}\times \vc{M}^{(3)}\times\vc{M}^{(4)}\times\vc{M}^{(5)}}=0 \]
is equivalent to test the variance components in Eq.(\ref{mee8}),    \[H_0:  \tau^{(1)} =\tau^{(2)} = \tau^{(3)} = \tau^{(1\times 2)} = \tau^{(1\times 3)}=\tau^{2\times 3} =\] \[\cdots=  \tau^{(1\times 2\times 3\times 5\times 5)}=0.\] 
It is known that  kernel matrices are not block-diagonal and   the parameter in variance component analysis are placed on the edge of the parameter space when the null hypothesis is true.  While the asymptotic distribution of a likelihood ratio test (LRT) statistic is  neither a chi-square distribution nor a  mixture chi-square distribution under the null hypothesis, we can use a score test statistic on the restricted likelihood \cite{Alam-18c, Liu-07}. We derive  score test statistic for a  generalized kernel machine approach  in multi-view datasets, Eq. (\ref{mee8}).  Assuming that  $\Theta (\gvc{\theta})= \sigma^2\vc{I} + \sum_{i=1}^5\tau^{(i)}\vc{K}^{(i)}+  \sum_{i=1}^5\sum_{j=i}^5\ \tau^{i\times j}\vc{K}^{(i\times j)}+\cdots+ \tau^{1\times 2\times 3\times 4\times 5}\vc{K}^{(1\times2\times 3 \times 4\times 5)}$, where $\gvc{\theta} = (\sigma^2, \tau^{(1)}, \tau^{(2)},  \tau^{(3)}, \tau^{1\times 2}, \tau^{2\times 3}, \cdots, \tau^{1\times 2\times 3 \times 4\times 5})$.  We can write the restricted log-likelihood function of Eq. (\ref{mee8}) as follows
\begin{eqnarray}
\label{mee9}
\ell_R (\gvc{\theta})& =& -\frac{1}{2}ln (|\Theta (\gvc{\theta})|)- \frac{1}{2} ln (|\vc{1}^T \vc{\Theta}^{-1} (\gvc{\theta}) \vc{1}^T |)\nonumber\\ &-& \frac{1}{2} (\vc{y} - \vc{X}\gvc{\hat{\beta}})^T\gvc{\Theta}^{-1} (\gvc{\theta}) \vc{1}^T (\vc{y} - \vc{X}\gvc{\beta})\end{eqnarray}
 Using  the partial derivative  of Eq. (\ref{mee9}) with respect to each variance component.  We derive the estimate of the variance components  and  the score test statistic which is defined as 
\begin{eqnarray}
\label{overall}
S(\sigma_0^2)= \frac{1}{2\sigma^2_0} (\vc{y}- \vc{X}\hat{\gvc{\beta}})^T \vc{K}(\vc{y}- \vc{X}\hat{\gvc{\beta}})
\end{eqnarray}
where $\vc{K}= \vc{K}^{(1)}+ \cdots + \vc{K}^{(1\times 2)}+\cdots+ \vc{K}^{(1\times2\times 3\times 4\times 5)}$ and  $\hat{\gvc{\beta}}$ is the maximum likelihood estimator (MLE) of the logistic regression coefficient under the null model $\vc{y}= \vc{X}\gvc{\beta}+\epsilon_0$, $\sigma_0^2$ is the variance of $\epsilon_0$. $S(\sigma_0^2)$ is quadratic function of the variable $\vc{y}$, which follows a weighted mixture of the chi-square distribution under the null hypothesis.  By the Satterthwaite method, we can approximate the distribution of $S(\sigma_0^2)$ to a scaled chi-square distribution, ($S(\sigma_0^2)\sim \gamma \chi^2_\nu$). To estimate the  the scale parameter $\gamma$ and the degrees of freedom $\nu$  we can  we can use  the method of moments on the mean ($\gamma\nu$) and variance ($2\gamma^2\nu$) of the test statistic and we obtain
 $\hat{\gamma}=\frac{\rm{Var}[S(\sigma_0^2)]}{2\rm{E}[S(\sigma_0^2)]}$ and $\hat{\nu}=\frac{2\rm{E}[S(\sigma_0^2)^2]}{\rm{Var}[S(\sigma_0^2)]} $. 
Finally, to get the $p-$ value of an experimental  score statistic  $S(\hat{\sigma}_0^2)$  we use the  scaled chi-square distribution $\hat{\gamma}\chi^2_{\hat{\nu}}$.

 \subsubsection{Testing  marginal effects}
  By using  the respective kernel matrix, we can test the marginal effect for each view data. We known that the testing marginal  effect
 \[H_0:  h_{\vc{M}^{(\ell)}}= 0, \ell =1,2,3,4,5.\]
is equivalent to test the variance components in Eq.(\ref{mee8}),    \[H_0:  \tau^{(\ell)} =0.\] 
 The test statistic is 
\begin{eqnarray}
S(\tau^{(\ell)})=  (\vc{y}- \vc{X}\hat{\gvc{\beta}})^T \vc{K}^{(\ell)}(\vc{y}- \vc{X}\hat{\gvc{\beta}}) \nonumber \\
\ell=1,2,3,4,5 \nonumber
\end{eqnarray}
\vspace{2mm}
 This is same as the sequence kernel association test (SKAT) \cite{Wu-11}.
 \subsubsection{ Testing  interaction effect}
   By considering no marginal effects, i.e., $\tau^{(\ell)} =0, \,\ell= 1, \cdots, 5$,  we can also test the interaction effects. 
 We know that the testing interaction  effect
 \[H_0:  h_{\vc{M}^{(\ell\times \xi)}}= 0, \ell < \xi =1, 2, 3, 4, 5\]
is equivalent to test the variance components in Eq.(\ref{mee8}),    \[H_0:  \tau^{(\ell\times \xi)} =0.\] 
 The test statistic is 
\begin{eqnarray}
S(\tau^{(\ell\times \xi)})=  (\vc{y}- \vc{X}\hat{\gvc{\beta}})^T \vc{K}^{(\ell\times \xi)}(\vc{y}- \vc{X}\hat{\gvc{\beta}}) \nonumber \\
\ell< \xi=1,2,3,4,5 \nonumber
\end{eqnarray}
\vspace{2mm}
 Also by considering no marginal effects, i.e., $\tau^{(\ell)} =0, \,\ell= 1, \cdots, 5$, and all second order interactions are zero $\tau^{(\ell\times \xi)} =0,$  we can also test the 3rd order interaction effects. 
\[H_0:  \tau^{(\ell\times \xi \times  \zeta )} =0.\] 
 The 3rd order test statistic is 
\begin{eqnarray}
S(\tau^{(\ell\times \xi\times  \zeta )})=  (\vc{y}- \vc{X}\hat{\gvc{\beta}})^T \vc{K}^{(\ell\times \xi \times \zeta)}(\vc{y}- \vc{X}\hat{\gvc{\beta}}) \nonumber \\
\ell< \xi< \zeta=1,2,3,4,5 \nonumber
\end{eqnarray}
 and so on. 
\subsubsection{Testing  composite  effects}
If lower order effects are statistically significant,  we want to test higher order effects is called composite hypothesis testing.  To test the 3rd order interaction effect, we mansion that  testing the null hypothesis  
$H_0: h_{\vc{M}^{(1)}\times \vc{M}^{(2)}\times \vc{M}^{(3)}\times \vc{M}^{(4)}\times \vc{M}^{(5)}} (\cdot)=0$ 
is equivalent to testing the variance component: $H_0: \tau^{1\times 2\times 3 \times 4\times 5}=0$. Let $\Sigma= \sigma^2\vc{I} + \tau^{(1)}\vc{K}^{(1)}+ \cdots + \tau^{1\times 2}\vc{K}^{(1\times 2)}+ \cdots + \tau^{ 2\times 3\times 4\times 5}\vc{K}^{(2\times 3\times 4 \times 5)}$, and all $\tau$,  and   $\sigma^2$  are model parameters under the null model.We formulate a test statistic
\begin{eqnarray}
\label{hin}
S_I(\tilde{\theta})= \frac{1}{2\sigma^2_0} \vc{y}^T \vc{B}_I \vc{K}^{(1\times2\times 3 \times 4 \times 5)}\vc{B}_I\vc{y},\end{eqnarray}
 where $\tilde{\theta} = (\sigma^2, \tau^{(1)}, \tau^{(2)},  \tau^{(3)}, \tau^{(1\times 2)}, \tau^{(2\times 3)})$, and  $\vc{B}_I= \Sigma^{-1}- \Sigma^{-1}   \vc{X}(\vc{X}^T  \Sigma^{-1}  \vc{X})^{-1} \vc{X}^T\Sigma^{-1}$ is the projection matrix under the null hypothesis.
 
Similar to the overall effect test,  we can use the  Satterthwaite method to  approximated the distribution of   higher order intersection test statistic $S_I(\tilde{\theta})$ by a scaled chi-square distribution  with scaled   $\gamma_I$ and degree of freedom $\nu_I$ i.e., $S(\tilde{\theta})\sim \gamma_I \chi^2_{\nu_I}$.
 The scaled parameter and degree of freedom  are estimated by MOM, $\hat{\gamma}_I=\frac{\rm{Var}[S_I(\tilde{\theta})]}{2\rm{E}[S_I(\tilde{\theta})]}$ and $\hat{\nu}_I=\frac{2\rm{E}[S_I(\tilde{\theta})]}{\rm{Var}[S_I(\tilde{\theta})]}$, receptively.
 In practices, the unknown  model parameters   are estimated by their respective  ReML estimates under the null hypothesis.   Lastly, the $p-$ value of an  observed higher order interaction effect test  score statistic  $S_I(\tilde{\theta})$ is obtained   using the  scaled chi-square distribution $\hat{\gamma}_I\chi^2_{\hat{\nu_I}}$.
 and so on. 

\subsubsection{Kernel machine based model selection}
The result of kernel-based machine learning approaches is highly influenced by the kernel and its parameters. The model selection with a suitable kernel is essential in machine learning.  As a result, machine learning approaches suffer from weak selection of a suitable model.  Suppose $k:\mc{X}\times \mc{X}\to \mb{R}$ is a positive definite kernel.  A linear positive definite kernel, $k(\vc{X}_i, \vc{X}_j)= \vc{X}_i^T\vc{X}_J$, where  $X, \tilde{X} \in \mc{X}$,    on $\mb{R}$  can define the similarity measure in the Euclidean space.  This liner kernel suffers from  more complexity in the function space for high-dimensional datasets. Instead of linear kernel, we can use a polynomial kernel to make it possible to  use higher order correlations among data points. A polynomial kernel, $k(\vc{X}_i,\vc{X}_j)= (X_i^T \vc{X}_j+c )^d, \, (c\geq 0, d\in \mb{N})$ has two free parameters: c is a tradeoff between higher order and lower order in the polynomial and d is the degree of the polynomial. The dependency measure depends on the finite bounded degree. Another limitation of linear and polynomial kernel is boundedness, i.e., both  kernels are unbounded.   

From a finite  dimensional space the Gaussian  kernel can map   the input  space (data space)  into a infinite dimensional space \cite{Schlkof-book} and is  defined as:

\[
k(\vc{X}_i, \vc{X}_j )=e^{-\frac{1}{2\sigma^2}||X_i- X_j ||^2},\, (\sigma > 0).
\]

It has a free parameter to select but consist of numerous theoretical properties including  boundedness,  consistence,  universality, robustness, and so on \cite{Sriperumbudur-09}. To fix the free parameter, we can use the median of the pairwise distance\cite{Gretton-08, Song-12}.

  For a multi-omics study, to  captures the pairwise  similarity across a number of SNPs in each gene we can use a  identity-by-state (IBS) kernel (nonparametric function of the genotypes) \cite{Kwee-08}:  
\[ k(\vc{M}_i, \vc{M}_j)= 1- \frac{1}{2s}\sum_{b=1}^s| M_{ib}- M_{jb} |,\]
where $s$ is the number of SNP markers of the corresponding gene. It is  assumption free on the type of genetic interactions. Hence, it can capture any genetic effects on the phenotype. In this paper, we used the IBS  for the genetic dataset and  for all other datasets, respectively.

\section{Experiments}
\label{Sec:exp}
We conducted experiments on both the  Monte Carlo simulation and two real multi-view datasets: an imaging and genome datasets  from the adolescence brain development and   five omics datasets  from osteoporosis studies, respectively.  The goal is to select a  feature based on significant composite effects among the datasets. We considered   the Gaussian kernel (the median of the pairwise distance as the  bandwidth) For all datasets but the genetic dataset. For the genetic dataset, we considered the IBS kernel \cite{Gretton-08, Alam-18c}. The optimization of the proposed approach is based on Fisher's scoring algorithm (the ReML algorithm).  To that end, we follow the parameters setting as in \cite{Alam-18c} for the ReML algorithm applicable.  To overcome the potential of  a local minima, we considered a set of initial points in (0, 1) and picked the point which maximized the ReML algorithm.

We compare our simulation and real data results with three other methods: principal component SKAT, partial principal component regression (pPCAR), and full principal component regression (fPCAR)) as  logistic regression setting.  The SKAT approach is a flexible and computational in  GWAS \cite{Wu-11, Ionita-13}.   To identify an interaction effect of two genes, a principal component   analysis based approach has been proposed by \cite{Li-12}.  \cite{Alam-18c} has been considered  each  method as a simple regression setting. Here, we switched each method as  a logistic regression setting.
 
 \subsection{Simulation studies}
   To evaluate the performance of the proposed method, we do simulation studies.  To that end, we   use the following model to generate synthesized qualitative phenotype:

\begin{multline}
\label{exper1}
y_i = \rm{logit} [\vc{X}_i^T \beta + \sum_{\ell=1}^5 \alpha_1 h_{\vc{M}^{(\ell)}}(\vc{M}_{i}^{(\ell)}) +  \sum_{\ell < \xi} \alpha_2h_{\vc{M}^{(\ell)}, \vc{M}^{(\xi)}} (\vc{M}^{(\ell)}_{i}, \vc{M}^{(\xi)}_{i}) + \sum_{\ell < \xi< \zeta} \alpha_3h_{\vc{M}^{(\ell)}\times \vc{M}^{(\xi)}\times \vc{M}^{(\zeta)}} (\vc{M}^{(\ell)}_{i}, \vc{M}^{(\xi)}_{i},  \vc{M}^{(\zeta)}_{i}) +  \sum_{\ell < \xi< \zeta < \tau} \alpha_4h_{\vc{M}^{(\ell)}\times \vc{M}^{(\xi)}\times \vc{M}^{(\zeta)   } \times \vc{M}^{(\tau)}}(\vc{M}^{(\ell)}_{i}, \vc{M}^{(\xi)}_{i},  \vc{M}^{(\zeta)}_{i},  \vc{M}^{(\tau)}_{i} )+  \alpha_5 h_{\vc{M}^{(1)}\times \vc{M}^{(2)}\times \vc{M}^{(3)} \times \vc{M}^{(4)}, \times \vc{M}^{(5)}} (\vc{M}^{(1)}_{i}, \vc{M}^{(2)}_{i}, \cdots, \vc{M}^{(m)}_{i}) +\sigma \epsilon_i], \nonumber 
\end{multline}

where $\vc{X}_i$ is a vector of covariates (height and weight) with  an intercept of  $i$-th  subject ($i=1,2, \cdots, n $) and   $\beta$ is a coefficient vector, the random error, $\epsilon_i$, follows the Gaussian distribution with  zero mean  and unit variance and  $\sigma$ is the  standard deviation of the error and is fixed  to $10^{-02}$.  Here $M_i$'s    represent five different datasets.  Each function $h_M(M_i)$'s is  defined as in different non-linear  form.

In this  setting,  we simulated data under  different values of parameters $(\alpha_1, \alpha_2, \alpha_3, \alpha_4, \alpha_5)$ to evaluate the performance  of the test.  For example,  $\alpha_1= \alpha_2= \alpha_3=0 = \alpha_4= \alpha_5=0$  means all effects have  vanished and we  can  examine the false positive rate  for  the score test under  the  overall effect.  Similarly, we take into account the   main effects (2nd order interaction effects) but no higher order interaction effects to  evaluate the power of the score test.  To get  consist results, we   performed  $1000$ simulations for  each setting.

The simulation results of different methods can be found in Table~\ref{tab:simu1}.  In this table, we report the power of that  higher-order composite score test in different parameter settings. Overall, we made two observations.  First, we noticed that the false positive rate of the test for higher-order composite effects score test  was controlled by fixing the nominal $p-$value threshold to $0.05$.  All four methods have the same observations for the false positive rate. Second, when considering the power analysis ($\alpha_5\geq 0$), we observed that    the proposed method performs better than other methods and its power   exceeds $0.80$ (Table~\ref{tab:simu1}). On the other hand we observed that the state-of-the-art methods (pPCAR, fPCAR and SKAT) can significantly   overstate the false positive rates and result in significant loss of statistical  power.

\begin{figure*}
\begin{center}
\includegraphics[width=14cm, height=6cm]{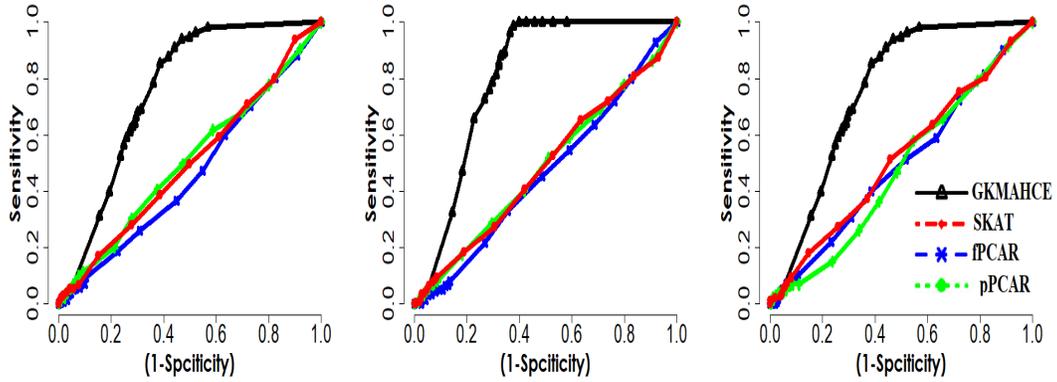}
\caption{The Receiver Operating Characteristics  of the kernel methods and relevant ones for higher-order composite effects detection with three sample sizes, $n\in \{100, 500, 1000\}$ for  all interrelated parameters ($\alpha_2 = \alpha_3 =  \alpha_4 = \alpha_5$) values are random but  the linear parameter is fixed to  $\alpha_1 =1$. }
\label{fig:ROC}
\end{center}
\end{figure*}

We also plot several visualization of   the ROC with  all interrelated parameters ($\alpha_2 = \alpha_3 =  \alpha_4 = \alpha_5$) values are random but the linear parameter is fixed to  $\alpha_1 =1$.  We allocate each number with probability $0.5$.  A random number is uniformly distributed  either in a range or at $0$.  In Figure~\ref{fig:ROC},  we look into the results  from the  receiver operating characteristic (ROC)  with three sample sizes, $n\in \{100, 500, 1000\}$.  The sensitivity is plotted against (1- specificity) for each $p$-value in the range of $0 - 1$ with a step size $0.0001$. The power gain of the proposed method relative to the alternative ones is evident in all situations.
 \begin{table}[!tbp]
\begin{center}
\caption {The power of  higher-order composite  score test  of the  proposed method (GJNAGCE), and state-of-the-art methods  using dimension reduction regression (e.g., pPCAR, fPCAR) and sequence kernel association test (SKAT)}
\begin{tabular}{|l|cccc|}
\hline
\rm{Parameters}&\multicolumn{4}{c|}{\rm{Simulation - I}}\tabularnewline
&\multicolumn{1}{c}{\rm{GKMAHCE}}&\multicolumn{3}{c|}{\rm{State-of-the-art methods}} \tabularnewline
&\multicolumn{1}{c}{}&\multicolumn{1}{c}{\rm{pPCAR}}&\multicolumn{1}{c}{\rm{fPCAR}} &\multicolumn{1}{c|}{\rm{SKAT}} \tabularnewline
\multicolumn{1}{|l|}{($\alpha_1$, $\alpha_2$, $\alpha_3$, $\alpha_4$, $\alpha_5$)}&&&&\tabularnewline
\hline
(0.1, 0, 0, 0, 0)&$0.059$&$0.048$&$0.0500$&$0.043$\tabularnewline
(0, 0, 0, 0, 0.1)&$0.841$&$0.044$&$0.053$&$0.042$\tabularnewline
(0, 0, 0, 0, 0.5)&$0.849$&$0.043$&$0.052$&$0.043$\tabularnewline
(0, 0, 0, 0, 1)&$0.828$&$0.047$&$0.052$&$0.045$\tabularnewline
(1, 1, 1, 1, 0.1)&$0.840$&$0.044$&$0.048$&$0.042$\tabularnewline
(1, 1, 1, 1, 0.5)&$0.836$&$0.049$&$0.048$&$0.043$\tabularnewline
(1, 1, 1, 1, 1)&$0.834$&$0.050$&$0.050$&$0.044$\tabularnewline
\hline
\end{tabular}
\label{tab:simu1}
\end{center}
\end{table}
 
 \subsection{Real data analysis}
We apply the proposed method to two real multi-view datasets including  an  fMRI imaging dataset and   five omics datasets from the adolescence brain development and osteoporosis studies, respectively. 
 
 \subsubsection{Application to adolescence brain development}
\begin{figure}
\begin{center}
\includegraphics[width=8cm, height=7cm]{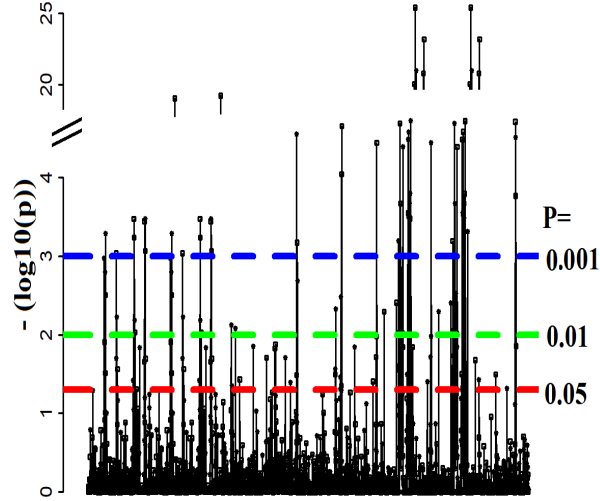}
\caption{The  plot of  $-\rm{log}_{10} (p)$ with  $3000$ triplets (gene, clinical measurement and region of interest.}
\label{fig:pval}
\end{center}
\end{figure}

 The Philadelphia Neurodevelopmental Cohort (PNC) is a research initiative  that focuses on characterizing brain and behavior interaction with genetics.  Sex differences in human brain are essential to understand their anatomical foundations in the brain \cite{Satterthwaite-46, Tunc-16}.    Our goal here is to select features using the test   composite effect  on   the  sex differences in cognition in  youth. To that end we consider three view datasets: genomic, non-genomic and fMRI imaging datasets. Table~\ref{tbl:pncdata} presents the configuration of the three  datasets along with phenotype. The gene consists  of genetic features (SNPs);  the non-genomic data   consists of  $116$ clinical measurements; and the  ROI consists of imaging features (voxels).  We consider each gene,  all non-genomic data and each ROI as a unique testing unit. To apply the proposed method, first we selected $231$ most significant genes out of  $13,150$ genes using the SKAT  approach and $241$ ROIs out of $264$ (since $23$ ROIs contain missing values). In total,  we get $55,671$ ($231 \times 1\times241)$ triplets to test. The overall test gives us $3000$ significant triplets ($p \leq 0.05$) out of $55,671$.  In Figure~\ref{fig:pval},  we  visualize the plot of $-(log10(p))$ for $3000$  triplets. The vertical solid, doted and double doted lines correspond to the p-values of $0.05, 0.01$, and $0.001$, respectively.  At p-values of $0.05$ and $0.001$ we  detected  $33$ and $14$ significant genes, as well as  $19$ and $12$  significant ROIs,  respectively.   

\begin{table}
\caption{The configuration of the three  data views of  adolescence brain development}
\begin{center}
\label{tbl:pncdata}
\scalebox{1}[1]{
 \begin{tabular}{l|ccc} \hline
 \rm{} & \rm{Genomic}& \rm{Non-genomic}& \rm{Imaging}\\ 
& \rm{SNP}& \rm{Clinical measurement}& \rm{Voxel}\\ \hline
  & \rm{95639}& \rm{116}& \rm{921429}\\
  \rm{Feature} &\multicolumn{1}{c}{\rm{Gene}}&&\multicolumn{1}{c}{\rm{ROI}}\\ 
   & \rm{13150}& & \rm{264}\\ \hline
   \rm{Subject} & \rm{863}& \rm{ 8719}& \rm{ 897} \\
   & \rm{common}& \multicolumn{2}{c}{\rm{798}}\\\hline
   Covariate & \multicolumn{1}{c}{\rm{Age}}\\\hline
   Phenotype  & \multicolumn{3}{c}{\rm{Gender}}\\
    &    \multicolumn{3}{c}{\rm{421 male and 377 female}}\\\hline 
\end{tabular}
}
\end{center}
 \end{table}
 
Table~\ref{tab:tgrg} presents the ReML estimates of all parameters, $\sigma^2$, $\tau^{(1)}$,  $\tau^{(2)}$, $\tau^{(3)}$,   $\tau^{(1\times 2)}$,  $\tau^{(1\times 3)}$,  $\tau^{(2\times 3)}$,  $\tau^{(1\times 2\times 3)}$ and  the $p$-values for both the proposed and SKAT methods for  each of the  $10$ triplets.  By the proposed method, these  $10$ triplets  were identified to have significant interactions at a level of  $p \leq  1.0 \times 10^{-7}$. At this  $p$-value, we observe $5$  unique genes ({\bf RPA3OS, ESR1, WWOX,  RGS7, PALLD}) and   $6$ ROIs  ({\bf  CFL.L}: Left cerebrum frontal lobe;  {\bf CSL.R}: Right cerebrum sub lobar; {\bf CLL.R}: Right cerebrum limbic lobe; {\bf CSS.R}: Right cerebrum cub-lobar;  {\bf CFL.R}: Right cerebrum frontal lobe), that may have  significant effects  of  sex  on adolescence brain development \cite{Baird-07, Gur-02}.

\begin{figure*}
\begin{center}
\includegraphics[width=16cm, height=14cm]{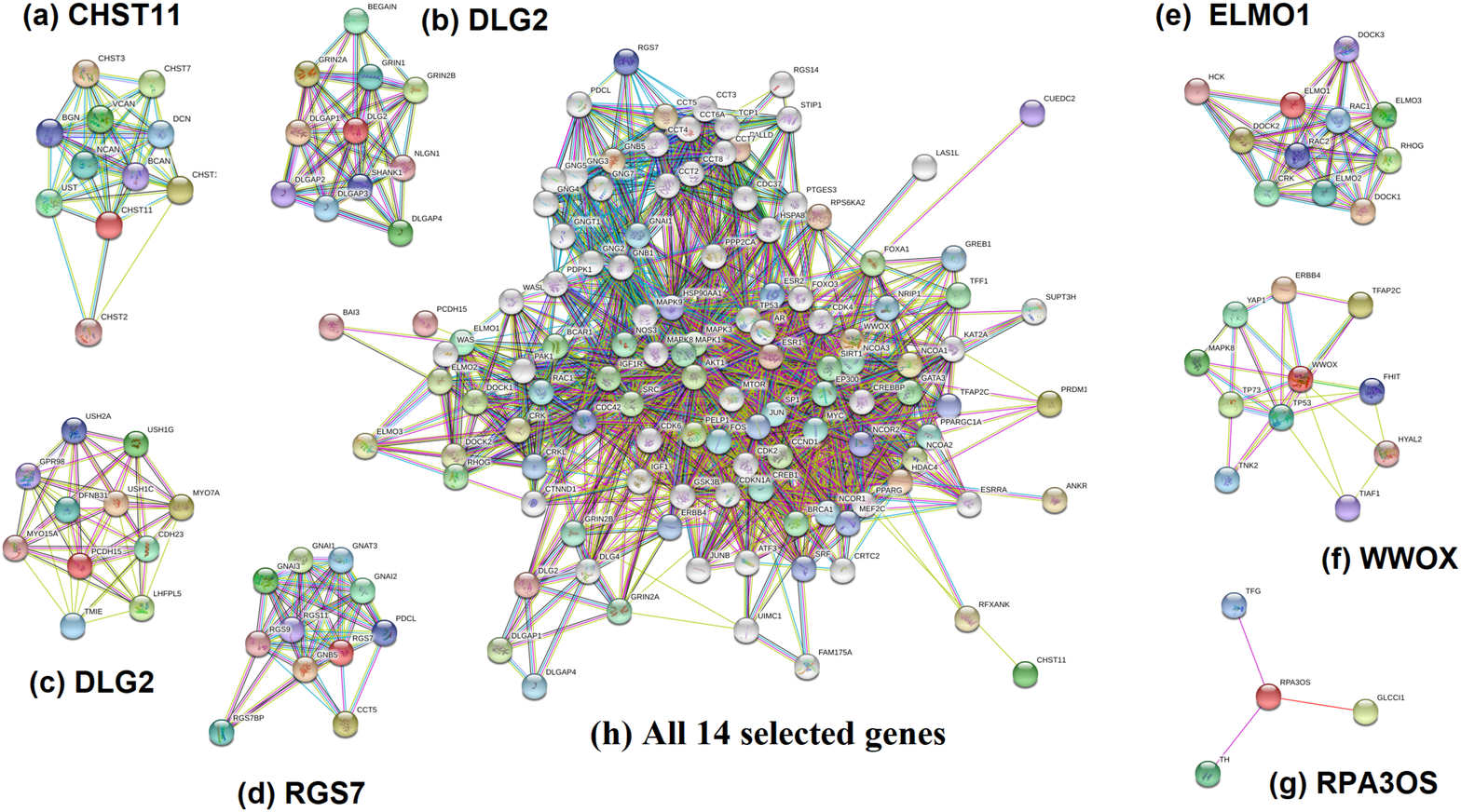}
 \caption{Network analysis of 7 individual (for example) selected genes ((a), (b), (c), (d), (e), (f), and (g)) and all  $14$ selected genes with the proposed methods: the network has significantly more interactions than expected. The color saturation of the edges represents the confidence score of a functional association. The colored nodes, white nodes, empty nodes, and filled nodes are noted for query proteins and the first shell of interactors, second shell of interactors, proteins of unknown 3D structure, and some 3D structure is known or predicted, respectively.}
\label{fig:net3d}
\end{center}
\end{figure*}

To test whether genes interact with each other, we  constructed a gene-gene interaction network analysis using STRING which imports gene/protein association knowledge from databases on both physical interactions and curated biological pathways \cite{STRING-15}. In STRING (e.g. STRING 11.0), the simple interaction unit is the functional relationship between two genes. This relationship can play a role to a common biological purpose. Figure~\ref{fig:net3d}  illustrates the gene-gene network based on the protein interactions with the selected $14$ genes ( including 7 individual selected genes). In this figure, the color saturation of the edges represents the confidence score of a functional association. Further network analysis shows  that the number of nodes, number of edges, expected number of edges, average node degree, clustering coefficient, $p$-values are $119$, $1478$,  $ 710$,  $24$, $0.654$ for $p\leq 1.0\times 10^{-16}$, respectively. This analysis confirmed that the selected genes had significantly more interactions than expected and  indicates that these genes may function in a concerted  manner.

\begin{table*}
\begin{center}
 \caption {The selected significant genes  and  ROIs using the proposed method and  the SKAT. The $p-$value was set to be $10^{-7}$ (CM: clinical measurement)}
\scalebox{0.6}[.6]{
\begin{tabular}{lcccccccccccccc}
\hline
&&&\multicolumn{10}{c}{\rm{GKMAHCE}}&\multicolumn{1}{c}{\rm{SKAT}}\tabularnewline
\multicolumn{1}{l}{}&\multicolumn{1}{c}{Genetics}&\multicolumn{1}{c}{Imaging}&\multicolumn{1}{c}{$\sigma^2$}&\multicolumn{1}{c}{$\tau^{(1)}$}&\multicolumn{1}{c}{$\tau^{(2)}$}&\multicolumn{1}{c}{$\tau^{(3)}$}&\multicolumn{1}{c}{$\tau^{1\times 2}$}&\multicolumn{1}{c}{$\tau^{1\times 3}$}&\multicolumn{1}{c}{$\tau^{2\times 3}$}&\multicolumn{1}{c}{$\tau^{1\times 2\times 3}$} &\multicolumn{1}{c}{\rm{OVA}} &\multicolumn{1}{c}{\rm{HOI}} &\multicolumn{1}{c}{\rm{HOI}}\tabularnewline\hline
&${\bf RPA3OS}$&$ {\bf CFL.L} $&$0.2001556$&$0.0737912$&$9.50E-06$&$0.0013817$&$0$&$0.0108441$&$6.19E-05$&$0.004694$&$0.0282793$&$0.000000$&$0.000663$\tabularnewline
&${\bf RPA3OS}$&${\bf CSL.R}$&$	0.23461$&$	0.1279504$&$	0.0026201$&$	0.001625$&$	0.0518874$&$	0.0009714$&$	2.40E-05$&$	0.0029383$&$	0.0122709$&$	0.000000$&$0.000663$\tabularnewline
CM&${\bf RPA3OS}$&$	{\bf CLL.R}$&$	0.159874$&$	0.2102631$&$	0.0006315$&$	0.0018936$&$0.1320889$&$	0.0027408$&$	0.0002114$&$	0.003986$&$	0.0335472$&$0.000000$&$0.000663$\tabularnewline
&${\bf RPA3OS}$&${\bf CSS.R}$&$	1$&$	7.71E-05$&$	0$&$	1.10E-05$&$	1.27E-05$&$	1.27E-05$&$	1.00E-05$&$	0.01$&$	0.0538638$&$	0.000000$&$0.00066$\tabularnewline
&${\bf WWOX}$&$	{\bf CFL.R}$&$	0.1132409$&$	0.125048$&$	0.0023859$&$	0.0024528$&$	0.0916626$&$	0.0027749$&$	4.37E-05$&$	0.0067442$&$	0.0036776$&$	0.000000$&$0.008035$\tabularnewline
&${\bf WWOX}$&$ {\bf CSL.R}$&$	0.3251332$&$	0.1203552$&$	0.0026668$&$	0.0024641$&$	0.1994204$&$	0.0011322$&$	0$&$	0.002699$&$	0.0108873$&$	0.000000$&$0.00803$\tabularnewline
&${\bf RPA3OS}$&$ {\bf	CLL.R}$&$	0.4310511$&$	0.0823752$&$	0.000000$&$0.0007619$&$	0.0507566$&$	0.0042518$&$	0.0003107$&$	0.01$&$	0.0335472$&$	0.000000 $&$0.000663$\tabularnewline
&${\bf RGS7}	$&$ {\bf CSL.R}$&$	0.1646353$&$	0.0820811$&$	0.0030577$&$	0.0021177$&$	0.0331062$&$	0.003288$&$	0.0001511$&$	0.0025276$&$	0.0116451$&$	0.000000$&$0.004248$\tabularnewline
&${\bf PALLD}$&${\bf CSL.R}$&$0.3705903$&$	0.079278$&$	0.0152793$&$	0.0014787$&$	0.0584442$&$	0.0004965$&$	9.79E-05$&$	0.0021067$&$	0.0107015$&$	1.00E-07$&$0.004536$\tabularnewline
\hline
\end{tabular}
}
\label{tab:tgrg}
\end{center}
\end{table*}

 For investigating  classification accuracy  of the proposed method, the  genome, non-genome, fMRI imaging datasets and only selected features via  the proposed method are used to classify the sex-related brain differences followed by the three classifiers: the K-nearest neighbors algorithm (KNN) and the support vector machine (SVM)  (the linear kernel and the Gaussian kernel). For the proposed approach, we considered triplets that have significant  effects on the gender: $14$ genes and $12$ ROIs only (from  Table~\ref{tab:tgrg}). The classifier is used to  distinguish  the adolescence brain development between the female and  the male. Table~\ref{tb:cerror3d} presents the train classification error with different datasets and the selected features only. These results  demonstrated that the proposed method is a better tool for feature selection.

\begin{table}
 \begin{center}
\caption {The training classification error of discriminating low BMD subjects  from high BMD with  the K-nearest neighbors (K-NN) algorithm  and  the support vector  machine (linear kernel and  Gaussian kernel) }\vspace{2mm}
\label{tb:cerror3d}
 \begin{tabular}{lcccccccccc} \hline
\rm{Dataset}&{\rm{ K-NN algorithm }}&{\rm{Linear kernel}} &{\rm{Gaussian kernel}}\tabularnewline\hline
{\bf Genome}& 23.06& $0.02$&$ 0.03$\tabularnewline\hline
{\bf Imaging}&$21.06$ & $0.05$& $0.10$\tabularnewline\hline
{\bf Non-genome}& $23.18$& $0.02$&$ 0.03$\tabularnewline\hline
{\bf Only selected features}& 21.05& $0.02$& 0.07\tabularnewline\hline
\end{tabular}
\end{center}
\end{table}

\subsubsection{Application to osteoporosis study}
Osteoporosis is a bone disorder which is largely attributable to  the increased bone resorption and/or decreased bone formation by osteoclasts and osteoblasts, respectively. We apply the proposed method to a multi-omics datasets from an osteoporosis study \cite{Yu-20, Qiu-20, Alam-19a}.  This dataset included genome,   epigenome,  transcriptome,  metabolome, and the lipidome data from  $57$ Caucasian females  with high BMD and $51$ with low BMD.
Table~\ref{tbl:ostdata} presents the configuration of the five  datasets along with phenotype. The original genomic data consists of  $3, 997, 535$ SNPs which annotated to 25, 442 genes. The  epigenome data consists of $46,690$ methylations which annotated to  $4,676$ genes. This  methylation analysis focus on gene level/sliding windows. The  transcriptome  data  has $22,682$ genes expression profiles. The  metabolome data contains of $28$ chemical taxonomies including $291$ metabolic profiles. The  lipidome data covers $7$ chemical taxonomies including $56$  lipid profiles. 

\begin{table}
\caption{The configuration of the five  data views of the  osteoporosis study (CT: chemical taxonomy)}
\begin{center}
\label{tbl:ostdata}
\scalebox{0.8}[.8]{
 \begin{tabular}{l|ccccc} \hline
 \rm{} & \rm{Genome}& \rm{Epigenome}& \rm{Transcriptome}& \rm{Metabolome}& \rm{Lipidome} \\ 
& \rm{SNP}& \rm{Methylation}& \rm{RNA-sequencing}& \rm{ profiling}& \rm{profiling} \\ \hline
  & \rm{3997535}& \rm{46690}& \rm{}& \rm{291}& \rm{56} \\
  \rm{Feature} &\multicolumn{3}{c}{\rm{Gene}}&\multicolumn{2}{c}{\rm{CT  class}}\\ 
   & \rm{25442}& \rm{4676}& \rm{22682}& \rm{27}& \rm{7} \\ \hline
   \rm{Subject} & \rm{129}& \rm{128}& \rm{128}& \rm{136}& \rm{136} \\
   & \rm{common}& \multicolumn{4}{c}{\rm{108}}\\\hline
   Covariate & \multicolumn{5}{c}{\rm{Age,~ Height\, and \,Weight}}\\\hline
   Phenotype  & \multicolumn{5}{c}{\rm{Bone mineral density (BMD)}}\\
    & \multicolumn{5}{c}{\rm{$51$ Low BMD  and $57$ High BMD}}\\\hline 
\end{tabular}}
\end{center}
 \end{table}

 \begin{figure}
\begin{center}
\includegraphics[width=8cm, height=8cm]{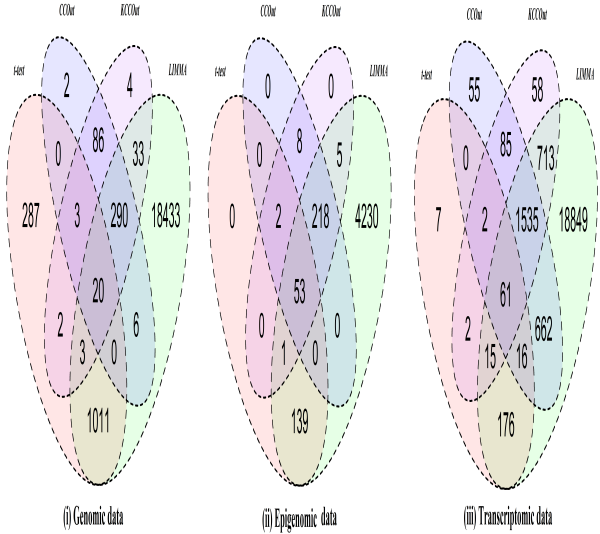}
 \caption{The   Venn diagram of the number of selected genes for  CCOut, LCCAOut, KCCAOut and LIMM  methods  with (a) the genome data; (b) the epigenome data  and (c) the transcriptome  data.}
\label{fig:venn3d}
\end{center}
\end{figure}

 \begin{figure}
  \begin{center}
\includegraphics[width=8cm,height=7cm]{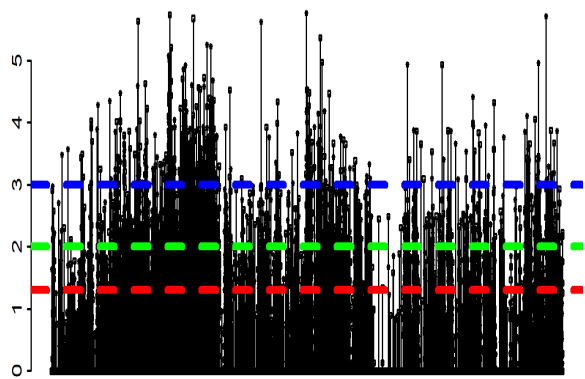}
 \caption{  The plot of $-log_{10}(p_{adj)}$ with $281,993$ quinlets (overall significance in the mode out of $12,2207,40$.}
 \label{fig:pval5d}
 \end{center}
\end{figure}

\begin{table}
\caption{The number of significant  genes (genome, epigenome and  transcriptome datasets) and chemical taxonomy  (Mmetabolome and  Lipidomoe datasets) of the proposed method  using adjusted p-values)}
\begin{center}
\label{tbl:ostdatap}
\scalebox{0.8}[1]{
 \begin{tabular}{lccc|cc} \hline
$P_{adj}$-values&\multicolumn{3}{c}{Gene}&\multicolumn{2}{|c}{Chemical taxonomy}\\ 
&Genome&Epigenome&Transcriptome& Metabolome& Lipidome  \\ \hline
$10^{-2}$& $19$&$51$ &$61$&$27$&$7$ \tabularnewline\hline
$10^{-3}$& $19$&$46$ &$56$&$26$&$7$ \tabularnewline\hline
$10^{-4}$& $19$&$40$ &$46$&$24$&$7$ \tabularnewline\hline
$10^{-5}$& $18$&$28$ &$35$&$22$&$7$ \tabularnewline\hline
$10^{-10}$& $17$&$35$ &$32$&$21$&$7$ \tabularnewline
\hline
\end{tabular}}
\end{center}
 \end{table}

\begin{table*}[ht]
\begin{center}
\caption {The selected  significant genes (genome, epigenome, and  transcriptome datasets) and chemical taxonomies (metabolome  and lipidome datasets)  using the proposed  at $p\leq 10^{-05}$}
\scalebox{0.6}[1]{
\begin{tabular}{lcccccccccccccccc}
\tabularnewline\hline
\rm{ {\bf Genome}}&{\bf ABCB8}&{\bf ACTBP13}& {\bf AP003122}& {\bf DOCK10}& {\bf EXOSC3P2}& {\bf FGF5}& {\bf GPLD1}&{\bf LINC00461}&{\bf CHGB}&{\bf NIPA2P5}&{\bf NIPAL1}\tabularnewline
&{\bf PANK1}&{\bf  RPL13AP3}  &{\bf  RPLP2}  &{\bf  SCRN2} &{\bf TRIM65} &{\bf ZNF114} &{\bf ZNF35}& & \tabularnewline \hline
\rm{ {\bf Epigenome}}
&{\bf A1CF}& {\bf  ALLC}& {\bf AP1G1}& {\bf ATP2B2}& {\bf BACH2}& {\bf CA13}&{\bf CCDC111}&{\bf DNER}& {\bf DYM}& {\bf FASTKD2} &{\bf GCNT7}\tabularnewline
& {\bf HCST}& {\bf INPP5F} & {\bf KIAA1305} & {\bf KNDC1} & {\bf NGTN} &
{\bf LGTN}& \bf {LMNB1}& {\bf LOC283871} & {\bf MAPKAPK2}& {\bf MAST4}& 
{\bf MDC1}\tabularnewline
& {\bf NOC2L}& {\bf NRL}& {\bf SLC44A3}& {\bf TRAIP}& {\bf UBE2J2}& {\bf VPS26A} & {\bf ZNF700} & &\tabularnewline\hline
 \rm{ {\bf Transcriptome}}&{\bf ABI1}&{\bf ACER3}&{\bf ADCY3}&{ \bf ALOX15B}&{\bf AUNIP}&{\bf BSND}& {\bf BTNL3}& {\bf CDKN1A}& {\bf C10orf105}& {\bf CROCC} &{\bf CSMD3} \tabularnewline
 & {\bf CTSLP2}& {\bf GALT} &{\bf  HEXDC} & {\bf HLA-DRA}   & {\bf HM13} & {\bf HMBOX1}   &{\bf IGHMBP2}& {\bf  KLRC1}  &{\bf LRRC8C} & {\bf MAGOH}  & {\bf MCTS2P}\tabularnewline
  & {\bf NEURL4} & {\bf NIN}  & {\bf NPHS1} & {\bf  PRPF38B} &  {\bf RALBP1} & {\bf  RNF17} & {\bf SERPINB6} &{\bf SLC38A9} &{\bf  SOX1} & {\bf TECTA } &{\bf TGFBR3L}   \tabularnewline
& {\bf USP17L17} & {\bf USP53}\tabularnewline\hline
\rm{ {\bf Metabolome}}&{\bf Azoles}&{\bf BENZ}&{\bf BSD}&{ \bf CAD}&{\bf CBS}&{\bf CXAD}& {\bf DIA}& {\bf FAE}
& {\bf FAL}&{\bf HAD} & {\bf HDMC}\tabularnewline
& {\bf IDA} & {\bf LAC} & {\bf PHE}& {\bf PLA} &{\bf PLL}   &{\bf PRNS}& {\bf PYRD}& {\bf SPHG} & {\bf STSD}  & {\bf TTRD} \tabularnewline
\hline
\rm{ {\bf Lipidome}}&{\bf AA}&{\bf  DGLA}&{\bf DHA}&{ \bf EA}&{\bf EPA}&{\bf ISTD}& LA & \tabularnewline\hline
\end{tabular}
}
\label{tab:ggg}
\end{center}
\end{table*}

\begin{figure*}[ht]
  \centering
  \begin{minipage}[b]{0.4\textwidth}
    \includegraphics[width=16cm, height=4cm]{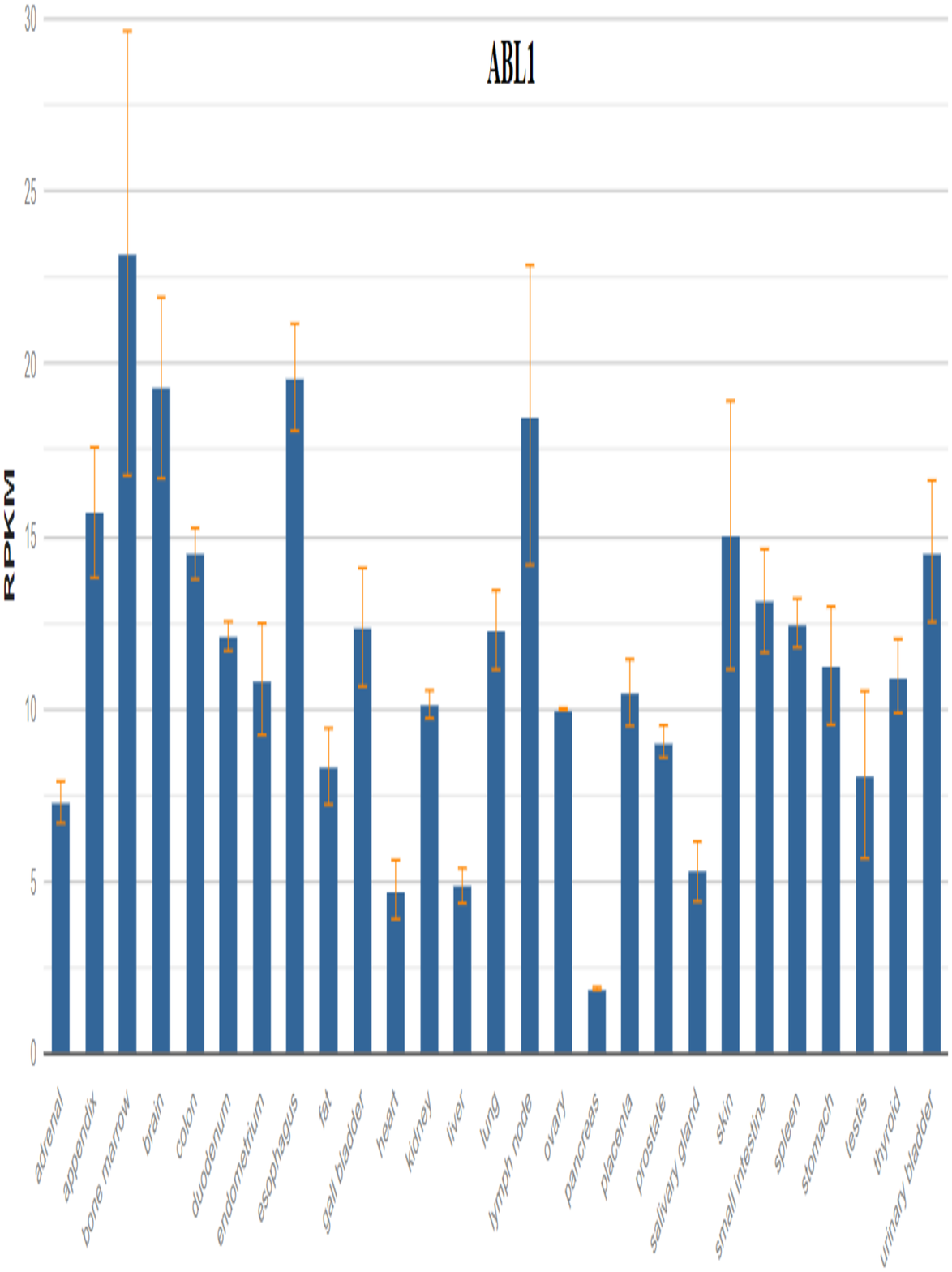}
  \end{minipage}
  \hfill
  \begin{minipage}[b]{0.4\textwidth}
      \end{minipage}
  \hfill
    \begin{minipage}[b]{0.6\textwidth}
    \includegraphics[width=16cm, height=4cm]{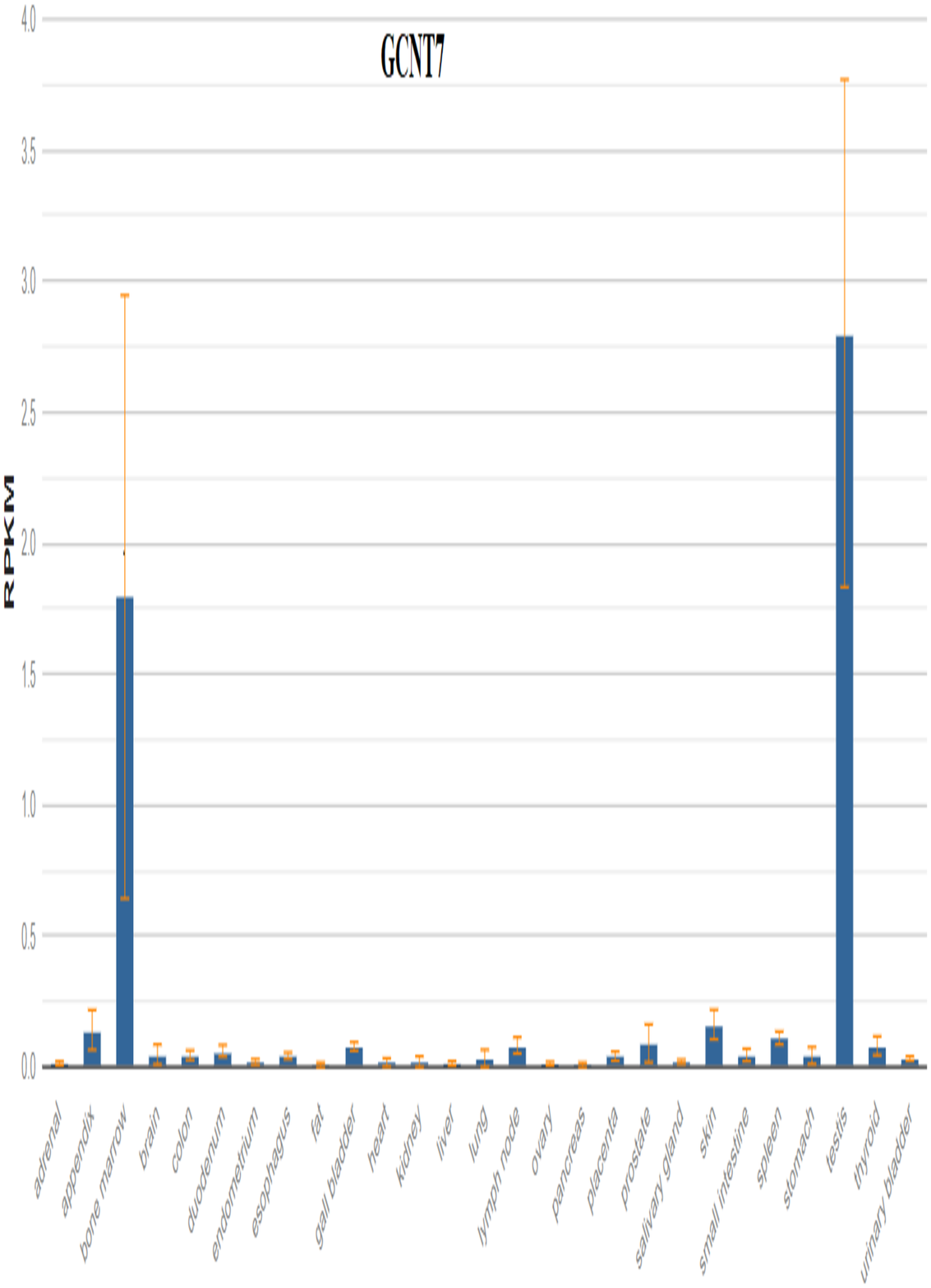}
     \end{minipage}
    \begin{minipage}[b]{0.4\textwidth}
  \end{minipage}
  \hfill
  \begin{minipage}[b]{0\textwidth}
  \end{minipage}
  \hfill
    \begin{minipage}[b]{\textwidth}
    \includegraphics[width=16cm, height=4cm]{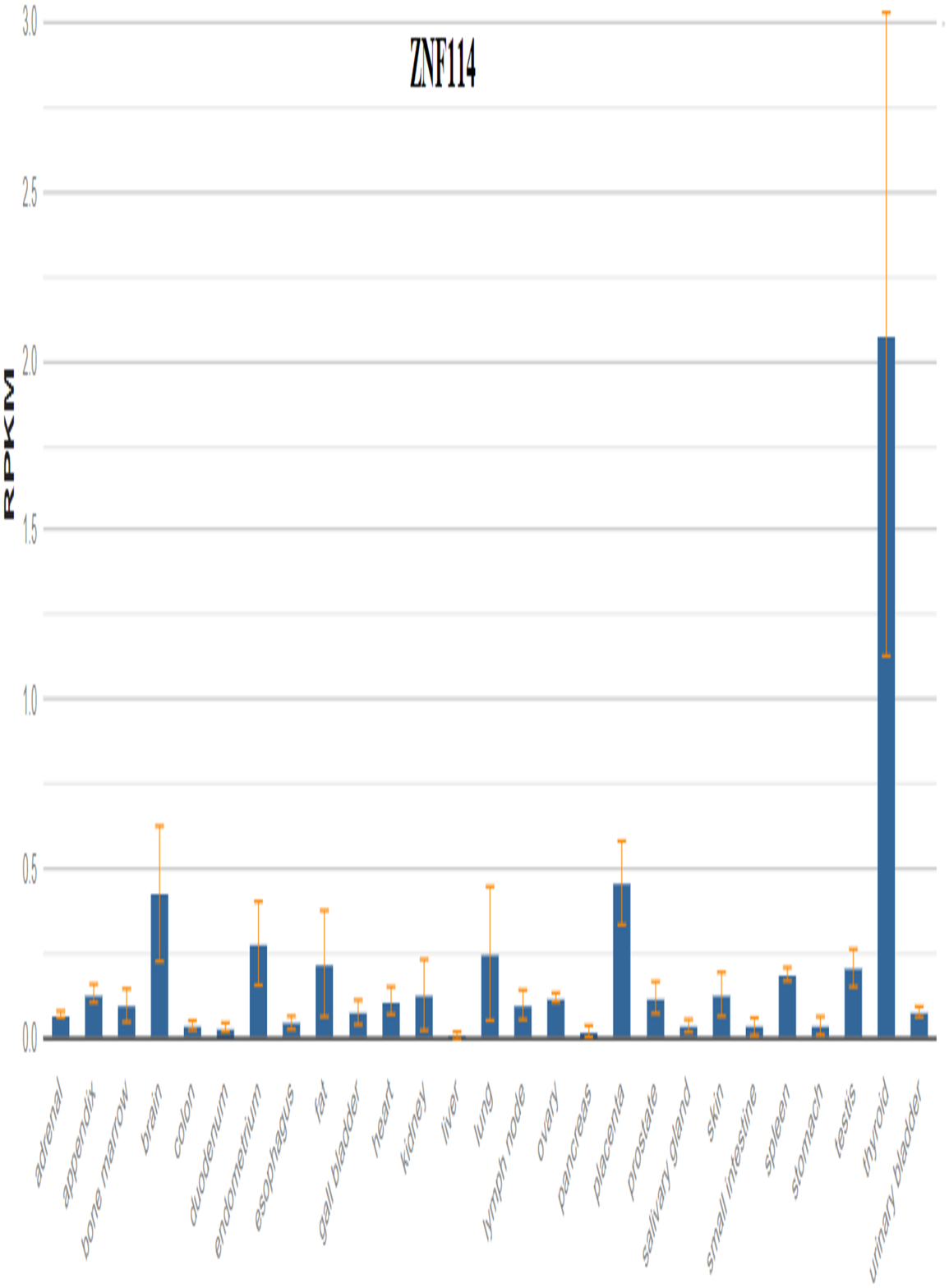}
     \end{minipage}
      \caption{The expression patterns of three selected genes (ABL1, GCNT7, ZNF114)  across $27$ different tissues. RPKM, reads per kilo base per million mapped}
      \label{fig:ncbi}
\end{figure*}

\begin{table*}
\caption{ The GeneHancer identifier, GeneHancer score, gene association score, total score, major-related diseases, and PubMed database (PMID)  of  6 selected   genes}
\begin{center}
\label{tbl:ostdatagene}
\scalebox{0.6}[1]{
 \begin{tabular}{l|c|c|c|c|c|c} \hline
\rm{Gene}  & \rm{ GeneHancer} & \rm{GeneHance}r& \rm{Association} &  \rm{Total}& \rm{Source}  &  \rm{Related} \\ 
 \rm{ID} &  \rm{ID} & \rm{Score} &  \rm{Score}&  \rm{ Score} & \rm{(Total No.):   PMID}  & \rm{ Disease} \\ \hline
\rm{ABL1} & \rm{GH09J130833}& $2.3$&$517.3$& $1214.2$& $(817)$:\,  $31025148$, $30704478$; $30666020;$ $29464484$; $29343719$& \rm{ Bone marrow;  Brain,
Endometrium; Lymphy node} \\ \hline
\rm {AP1G1} & \rm{ GH16J071807}& $2.1$&$520.0$& $1095.79$& $(103):$ \, $12773381$; $
11872161$; $24006255$; $16867982$; $28823958$& \rm{Bone marrow; Kidney; Brain, Testis; Thyroid} \\ \hline
\rm {GCNT7} & \rm{GH20J056524}& $0.4$&$500.7$& $221.23$& $(2)$: $25367360$; $25130324$& \rm {Bone marrow;  thyroid} \\ \hline
\rm {PANK1} & \rm{GH20J056524}& $2.2$&$506.7$& $1128.6$& $(21)$:  $12379284$; 
$11809413$; $14523052$;  $20833636$; $23343762$& \rm{Liver;
Kidney; Bone marrow; Thyroid} \\ \hline
\rm {ZNF35} & \rm{GH03J044648}& $1.9$&$500.7$& $975.9$& $(17)$: $2108922$;
$8477855$; $3380682$; $1572646$; $20186958$& \rm{Testis; Thyroid; Urinary bladder, Endometrium;  Brain, Bone marrow} \\ \hline
\rm {ZNF114} & \rm{
GH07J151024}& $2.4$&$513.1$& $1242.1$& $(8)$: $8467795$; $28065597$; $31515488$; $31515488$; $25416956$& \rm{Thyroid, Brain; Placenta} \\ \hline
\end{tabular}}
\end{center}
 \end{table*}

 \begin{figure*}[ht]
\begin{center}
\includegraphics[width=14cm, height=8cm]{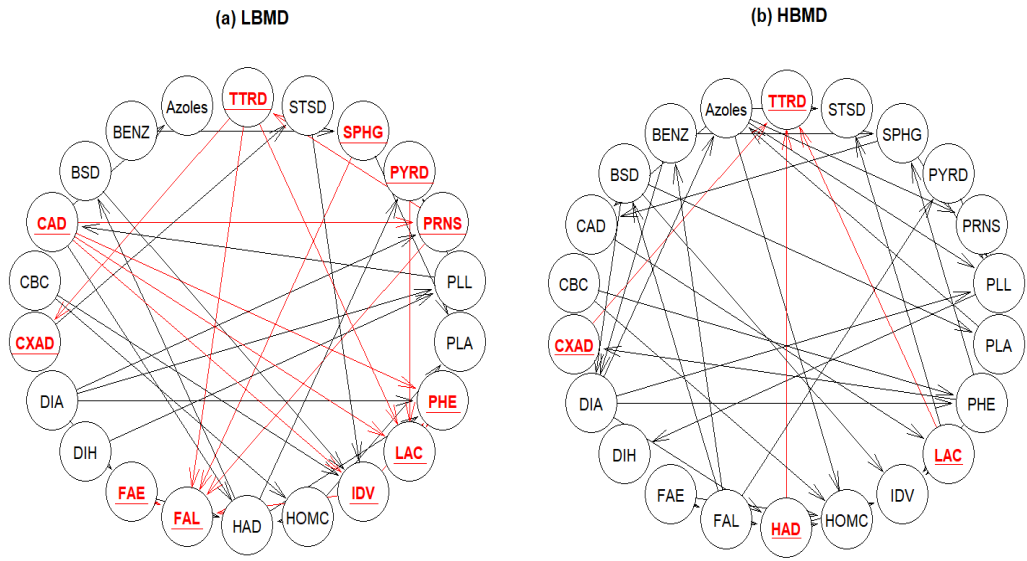}
 \caption{The   causal relation of the metabolome  dataset using  Hill-Climbing approach: (a) low BMD and (b) high BMD.}
\label{fig:cimb}
\end{center}
\end{figure*}

  \begin{figure}[ht]
\begin{center}
\includegraphics[width=8cm, height=7cm]{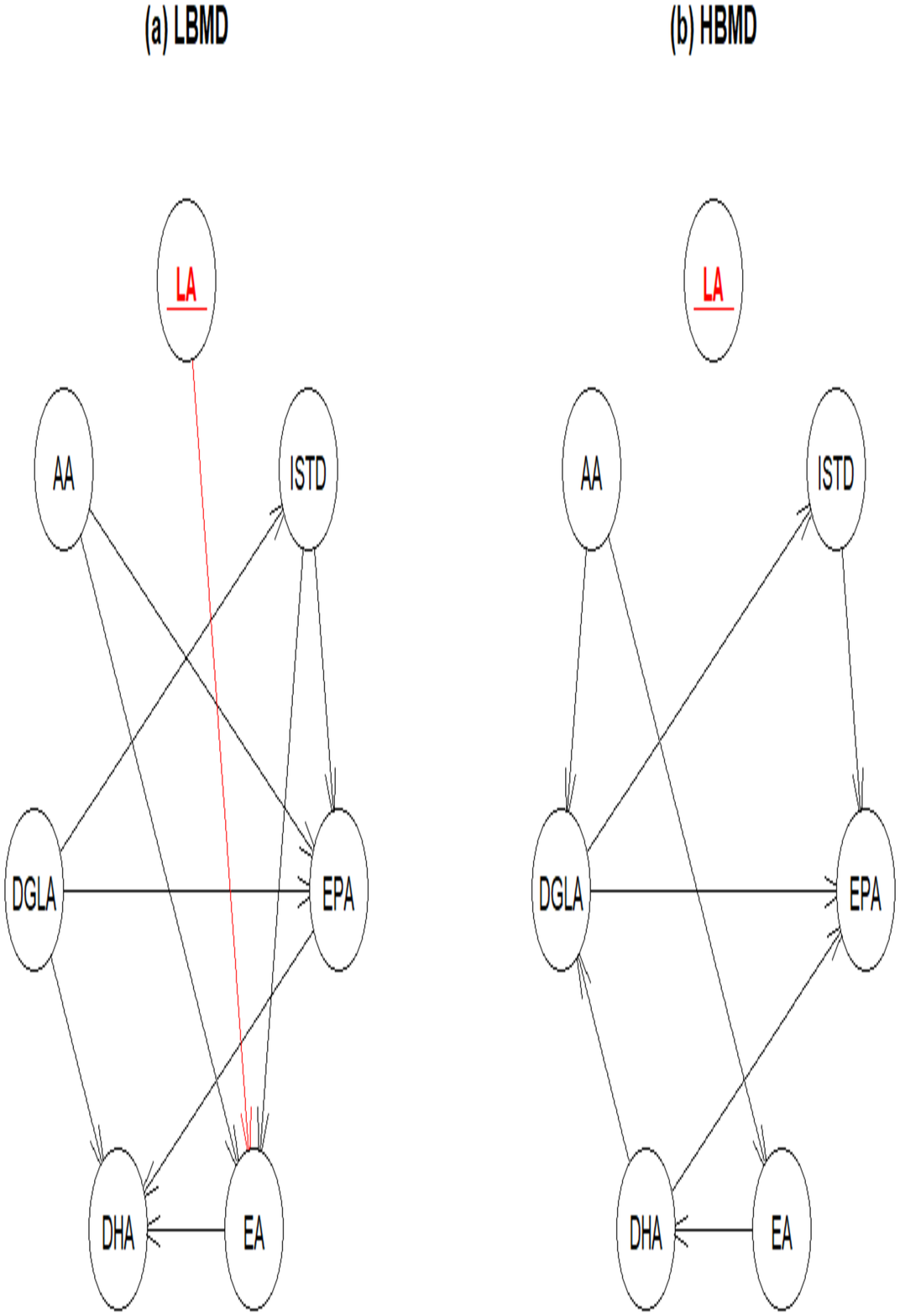}
 \caption{The   causal relation of the lipidomome  dataset using  Hill-Climbing approach: (a) low BMD and (b) high BMD.}
\label{fig:cilp}
\end{center}
\end{figure}

 To apply the proposed  method, we considered each gene of the  genome,  epigenome, and  transcriptome data and each chemical taxonomy of the metabolome and  lipidome data  as a single testing unit.  To make it feasible, we reduced the  dimensionality of the  genome, epigenome and  transriptome  data using four approaches including the  kernel based gene shaving   approach \cite{Alam-19a}. Figure~\ref{fig:venn3d} presents the Venn-diagram of the t-test, canonical correlation analysis based gene shaving (CCAOut),  kernel canonical correlation analysis based gene shaving  (KCCAOut) and Linear Models for Microarray based gene shaving (LIMMA) methods for three datasets: (a) genome data (b) epigenome data,  and (c)  transcriptome  data. From this figure, we observed that the number of disjointedly selected genes from CCOut, LCCAOut, KCCAOut and LIMMA methods for the genome data are $1,326$, $406$,  $442$, $19,796$;  for the epigenome data are  $193$, $281$, $287$, $4645$; and for the transcriptome  data are $279$, $2,416$, $2,529$, $22,027$, respectively.  All methods   selected  $20$, $53$ and $61$  common genes for  the genome,  epigenome and  transcriptome data, respectively. In addition,  we have   $28$ and $7$ chemical taxonomies of metabolome and  lipidome dataset, respectively. In total, we get $12220740$ ($201 \times 53\times 61\times 27\times 7)$  quinlets to test.   Figure~\ref{fig:pval5d},  shows the plot of $-(log10(p))$ for $12220740$  quinlets. The overall test gives us $281993$ significant quinlets ($p \leq 0.05$).   The vertical solid, doted and double doted lines correspond to the p-values of $0.05, 0.01$, and $0.001$, respectively.
 Table~\ref{tbl:ostdatap} presents the number of significant  genes  and chemical taxonomies using  different adjusted p-values of the proposed method.  A list of genes (genome, epigenome and transcriptome data) and list of chemical taxonomies (metabolome and   lipidome data ) are tabulated in Table~\ref{tab:ggg}.

We also extracted the gene-gene interaction networks using STRING: functional protein association networks. Figure S1,  (in the Supplementary material) shows the gene-gene networks based on the protein interactions among the selected ($81 18+28+35$) genes by  the proposed method of  the  genome, epigenome, and transcriptome datasets.  In this figure, the color saturation of the edges represents the confidence score of a functional association.  In addition, network analysis demonstrates that the number of nodes, number of edges, expected number of edges, average node degree, clustering coefficient, PPI  enrichment p-values are $93, 155, 121,  3.33,  0.39$ and $0.000188$, respectively.  This network analysis confirms that the selected genes have significantly more interactions than expected. It also indicates that the selected genes may function collaboratively.

To confirm the biological roles of the most prominent identified features, we accessed the biomedical and genomic information using  the national center for biotechnology information (NCBI, https://www.ncbi.nlm.nih.gov/) tools and the Human gene database (GeneCards: https://www.genecards.org/) \cite{Fishilevich-17}.  NCBI offers a wide variety of data analysis tools for manipulating, aligning, visualizing, and evaluating biological data. On the other hand, the primary goal of the GeneCards database is the unequivocal identification of enhancer elements and uncovering their connections to genes for understanding gene regulation and the molecular pathways. Recent studies have shown that the bone marrow, kidney, testis, and thyroid diseases have widespread systemic manifestations including their effects on developing osteoporosis. \cite{Kerai-18, Khairallah-18, Tuchendler-14, Willemse-14}. So, we hypothesized that the selected genes, with the proposed method, would provide more ubiquitous expression (reads per kilo base per million mapped reads (RPKM)) of bone marrow, kidney, testis, and thyroid tissue samples. To that end, we used NCBI’s tools.  Figure~\ref{fig:ncbi} illustrates the expression patterns of three selected genes (ABL1, GCNT7, ZNF144) across $27$ different tissues from 95 human individuals. These results show that these genes have remarkably high RPKM value for bone marrow, kidney, testis, and thyroid tissue samples. These tissue related diseases including their effects may play a role in the development of osteoporosis. To provide insight into the gene regulatory elements (promoters and enhancers) for 6 selected genes (ABL1, AP1G1, GCNT7, PANK1, ZNF35, ZNF144), we used the GeneCards database. Table~\ref{tbl:ostdatagene}  presents GeneHancer identifier, GeneHancer score, gene association score, total score, major-related diseases, and PubMed database. From this table, we observed that the selected genes have had a remarkable GeneHancer score, gene association score, total score, and literature review in the past studies. According to the disease annotation, the selected genes are highly associated with complex diseases, which are   higher risk of developing osteoporosis (the results of all selected genes  are in Supplementary material).

To further explore whether that the selected  chemical taxonomies of   metabolic and  lipidomic features  will have  unique relationships in the low BMD and high BMD groups. We infer a causal relationship with a Hill-Climbing approach among the select chemical taxonomies  in each group of  both datasets (metabolome  and lipidome).  The   causal relationship of the metabolome  and the lipidome datasets: (a) low BMD and (b) high BMD is illustrated in Figure~\ref{fig:cimb} and Figure~\ref{fig:cilp}, respectively.  In both datasets, we observed that the causal relationships are different in both the low BMD and high BMD for both datasets. For the metabolic data TTRD has an important impact on the low BMD but has an  negative impact on high BMD but   DA has  an important  impact on the  high BMD.  In case of the lipidome data, we observed that the LA has an impact on  the low BMD but does not have any impact in the high BMD.  By this observation, we  concluded that  the  selected  chemical taxonomies may contribute on BMD status but not in general.

Finally,  we  investigated  classification accuracy via only selected features  of the proposed method. To classify  the low BMD  subjects from the high BMD subjects  followed by the three classifiers: the KNN and the SVM  (the linear kernel and the Gaussian kernel). For the proposed approach, we considered triplets that have significant (from Table~\ref{tab:ggg}) effects on the BMD. The classifier is used to classify  the low BMD  subject from the high BMD subject. Table~\ref{tb:cerror5d}  presents the train classification error with different datasets and the selected features only. These results are  also demonstrating that the proposed method is a better tool for feature selection.

\begin{table}
 \begin{center}
\caption {The training classification error of discriminating low BMD subjects  from high BMD with  the K-nearest neighbors (K-NN) algorithm  and  the support vector  machine (linear kernel and Gaussian kernel) }\vspace{2mm}
\label{tb:cerror5d}
 \begin{tabular}{lcccccccccc} \hline
\rm{Dataset}&{\rm{ K-NN algorithm }}&{\rm{Linear kernel}} &{\rm{Gaussian kernel}}\tabularnewline\hline
{\bf Genome}& $30.28$ & $28.18$&$ 2.07$\tabularnewline\hline
{\bf Epigenome}& $26.01$ & $22.03$&$ 3.04$\tabularnewline\hline
{\bf Transcriptome}& $21.29$ & $0.00$&$ 0.05$\tabularnewline\hline
{\bf Metabolic}& $29.63$ & $0.00$&$ 0.22$\tabularnewline\hline
{\bf Lipidomic}& $25.92$ & $33.33$&$0.21$\tabularnewline\hline
{\bf Only selected feature}& $12.96$& $0.00$ &$0.08$\tabularnewline\hline
\end{tabular}
\end{center}
\end{table}

\section{Concluding remarks}
\label{Sec:dis}
   The technology of biomedical science  has  accelerated the cycle of  discovery. In this we  propose a novel generalized kernel machine approach to identify higher-order composite effects in multi-view biomedical datasets. This semi-parametric approach  considers multi-view data as prediction variables  to allow comprehensive modeling of complex disease trait.

The power of the proposed method is  further demonstrated by its application to both synthesized and two real multi-view datasets, i.e., adolescence brain development and osteoporosis study. In simulation studies, we have shown that the false positive rate of the test for higher-order composite effect  is  mitigated by fixing  the nominal p-value threshold along with other   state-of- the-art-methods.  From the power analysis, our proposed method not only performs better than other methods. The ROC curves also showed  the  gain of power by the  proposed method  in all scenarios. 

In adolescence brain development experiments, we found five unique genes and 6 ROIs along with the non-genomic factor that may have significant  gender effects  on adolescence brain development. In addition, by the gene-gene interaction networks, we confirmed that   the selected genes have significantly more interactions than expected. This result implies that the   gens and their functional interactions form the backbone of the cellular machinery.

 In the osteoporosis study, the network analysis confirmed that the selected genes of genome, epigenome, and transcriptome datasets also have  significantly more interactions than expected. It also indicates that the selected genes  may function in a collaborative effort. We found that  they  function in a concerted effort and have biological relevance to osteoporosis. Furthermore, using a causal relationship, we conclude that the selected chemical taxonomies (metabolome and lipidome) may contribute to development of BMD status.

The primary focus of our paper is to comprehensively characterize  complex  disease (e.g., adolescent brain development, or osteoporosis ). Collectively, we can use this approach to derive a statistic for testing the composite effect.  These studies explore novel genomic, epigenomic, transcriptomic and chemical mechanisms to identify corresponding factors. The significant triplet, quartet, quintet  of   extracted features suggest  that  these  effects  might highlight biological targets for drug development.  Extrapolating these findings  enable  an advanced understanding of normal cellular processes. Our proposed method  can be applied to the study of any disease models, where multi-view data analysis is commonly used.  

 We acknowledge that  the kernel machine is sensitive to contaminated data, even if bounded positive definite kernels are used.  Further work on robust  kernel methods including M-estimator based methods. They will allow us to uncover the rich, hidden connections  in  biological system, and may provide a more comprehensive modeling of complex traits.

\subsection*{Acknowledgments}
 This work was partially supported by grants from National Institutes of Health (NIH) [U19AG05537301, R01AR069055, P20GM109036, R01MH104680, R01AG061917], and the Edward G. Schlieder Endowment and the Drs. W. C. Tsai and P. T. Kung Endowment to Tulane University.


\bibliographystyle{plain}
\bibliography{Ref-2019}

\end{document}